\pdfoutput=1

\documentclass[11pt]{article}

\usepackage[preprint]{acl}

\usepackage{times}
\usepackage{latexsym}

\usepackage[T1]{fontenc}

\usepackage[utf8]{inputenc}

\usepackage{microtype}

\usepackage{inconsolata}

\usepackage{graphicx}

\usepackage{algorithm}
\usepackage{algpseudocode}
\usepackage{amsmath}
\usepackage{multirow}
\usepackage{array}
\usepackage{booktabs}
\usepackage{xspace}
\usepackage{amsfonts}
\usepackage{tcolorbox}
\usepackage{authblk}
\usepackage{soul}

\newcommand{\dashline}[2]{\noalign{\vspace{2pt}}\cline{#1-#2}\noalign{\vspace{2pt}}}

\algnewcommand{\LineComment}[1]{\State \textcolor{HarderGreen}{\(\triangleright\) #1}}

\algnewcommand{\EndComment}[1]{\textcolor{Green}{\Comment #1}}
\algrenewcommand{\alglinenumber}[1]{%
  \ifnum#1=1%
    \colorbox{figBlue}{\footnotesize#1:}%
  \else\ifnum#1=2%
    \colorbox{figGreen}{\footnotesize#1:}%
  \else\ifnum#1=5%
    \colorbox{EasyOrange}{\footnotesize#1:}%
  \else\ifnum#1=6%
    \colorbox{HardRed}{\footnotesize#1:}%
  \else\ifnum#1=7%
    \colorbox{figGreen}{\footnotesize#1:}%
  \else%
    \footnotesize#1:%
  \fi\fi\fi\fi\fi%
}

\definecolor{EasyOrange}{HTML}{FDCC8A}
\definecolor{HardRed}{HTML}{d7301f}
\definecolor{EasyGreen}{HTML}{17a324}
\definecolor{figGreen}{HTML}{d9ead3}
\definecolor{figBlue}{HTML}{c9daf8}
\definecolor{HarderGreen}{HTML}{0f6f18}
\definecolor{HarderRed}{HTML}{a32417}
\definecolor{HarderOrange}{HTML}{cc7f22}

\newcommand{\ie}{{\it i.e.}\xspace}
\newcommand{\eg}{{\it e.g.}\xspace}
\newcommand{\highlight}[1]{\colorbox{EasyOrange}{#1}}
\newcommand{\previoustok}[1]{\colorbox{figBlue}{\small{\st{#1}$\rightarrow$}}}

\newcommand{\ours}{\texttt{HARP}\xspace}
\newcommand{\oursfull}{Hesitation-Aware Reframed Forward Pass\xspace}
\newcommand{\oursfullletter}{\textbf{H}esitation-\textbf{A}ware \textbf{R}eframed Forward \textbf{P}ass\xspace}

\title{\ours: Hesitation-Aware Reframing in Transformer  Inference Pass}

\author[]{Romain Stora\"i}
\author[]{Seung-won Hwang\thanks{~~~Corresponding author.}}
\affil[]{Computer Science and Engineering, Seoul National University}
\affil[]{\texttt{\{romsto,seungwonh\}@snu.ac.kr}}

\begin{document}
\maketitle

\begin{abstract}
    This paper aims to improve the performance of large language models by addressing the 
    variable computational demands in inference steps, where some tokens require more computational resources than others.
    We present \ours, a simple modification to ``off-the-shelf'' Transformer forward pass.
    Drawing from hesitation and the framing effect in decision-making, \ours selectively applies additional computation when the model encounters uncertainty during token generation.
    Our method mimics human cognitive processes by pausing at difficult decision points and reframing inputs for a different perspective.
    Unlike other approaches, \ours is model-agnostic, training-free, and easy to implement.
    We evaluate our method across various downstream tasks and model sizes, demonstrating performance improvements up to +5.16\%.
    Notably, \ours achieves these gains while maintaining inference times twice faster than beam search.
    Simple and yet with significant gains, \ours provides insights into the potential of adaptive computation for enhancing the performance of Transformer-based language models.
\end{abstract}

\section{Introduction}
\label{sec:intro}

Causal language models based on the Transformer architecture \citep{vaswani2017attention} use a constant number of layer traversals to generate each new token.
While this architecture is beneficial to provide easy parallelization during training \citep{dehghani2019universaltransformers}, it may not fully leverage the model's full potential during inference, where tokens are generated sequentially.
Research in \emph{adaptive computation} (\eg \citealp{graves2017adaptivecomputationtimerecurrent}; \citealp{leviathan2023fast}; \citealp{elhoushi-etal-2024-layerskip}; \citealp{leviathan2024selectiveattentionimprovestransformer}) suggests that inference steps are not equally challenging, with some being ``harder'' and others ``easier.''
Intuitively, these more challenging tokens would benefit from additional computational resources to improve accuracy.
Unfortunately, the current Transformer architecture treats each token equally---regardless of its difficulty---potentially leading to imprecision and performance drops.

\begin{figure}[t]
    \centering
    \includegraphics[width=\linewidth]{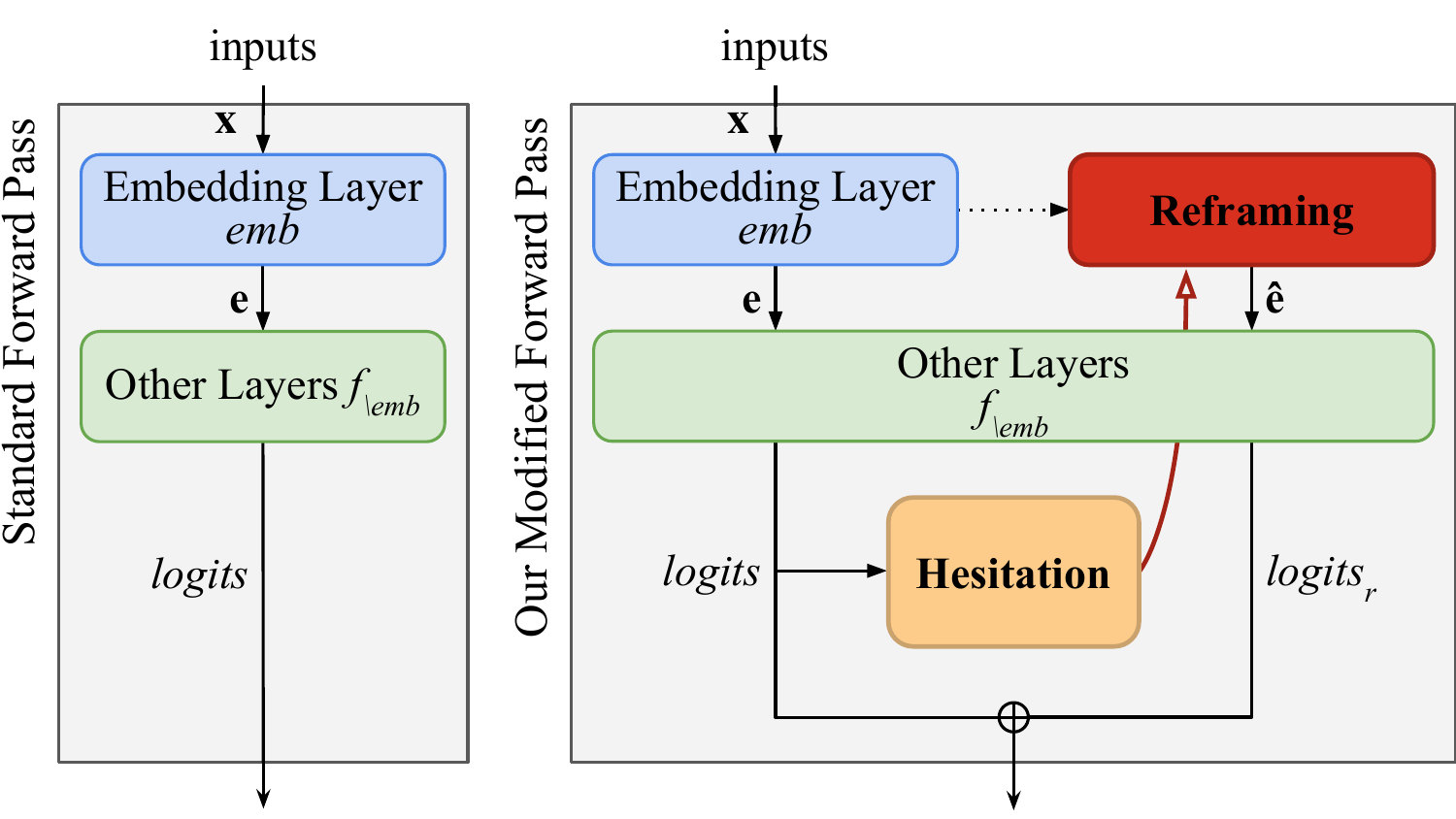}
    \caption{The left side represents the Transformer's vanilla forward pass, while the right side illustrates the modified forward pass, \ours, which selectively applies additional computation by reframing inputs when the model hesitates. This improves performance on ``harder'' tokens without the need for retraining.}
    \label{fig:bigpicture}
\end{figure}

To address this limitation, a trend in language models has been to simply scale up models in size \citep{brown2020gpt3}, allowing for more computation.
While the difficult tokens benefit from this scaling, it also leads to unnecessary overhead for easier tokens.
To tackle this uniform additional computation, Speculative Decoding \citep{leviathan2023fast, chen2023accelerating} employs a bigger model to verify and correct the tokens generated by the smaller model, acting as an external verifier.
Considering the smaller model as the original model, the larger model performs additional computations to ensure the quality of the generated tokens.
This enables the system to spend more computational resources on complex tokens while preserving efficiency for easier ones.
However, this approach requires the involvement of an external model.
Similarly, \citet{goyal2024think} add fixed pauses during inference, using ``pause tokens,'' to allow for additional computation on harder tokens.
While their method improves the generation of harder tokens, it requires full training and fine-tuning, and it still applies uniform additional computation when simpler tokens do not need the extra time.

We draw inspiration from human behaviors to allow models to perform additional computations for ``harder'' steps without relying on external models or requiring retraining.
Two cognitive effects stand out: the (1) \textbf{hesitation} and the (2) \textbf{framing effect}.
First, \emph{hesitation} reflects \underline{uncertainty} in decision-making.
Humans tend to pause and reconsider when faced with difficult decisions \citep{shenhav2013valueofcontrol} such that more effort is spent on complex inputs---aligning with the idea of ``harder'' tokens during inference.
Second, the \emph{framing effect} indicates that how information is presented can influence judgment and response \citep{kahneman_thinking_2012}.
It implies that a \underline{different representation} of the same input can lead to better outcomes.
In the following, we will refer to this another-view representation as ``\emph{reframing}'' the inputs. 

Building on these human-inspired concepts, we introduce \oursfull (\ours), a plug-and-play modification to the Transformer’s forward pass.
To illustrate, we summarize \ours on the right side of Figure \ref{fig:bigpicture}.
We begin with the standard forward pass, processing the inputs through the embedding layer and subsequent layers to produce initial logits.
We then evaluate the \emph{hesitation} of the model by computing its \underline{uncertainty} over the logits (detailed in Section \ref{sec:esti_unc}).
If the model is not hesitating, we directly output the initial logits.
However, when the model is in a hesitation state, we \emph{reframe} the inputs to infuse a \underline{different representation} (explained in Section \ref{sec:reframing}).
To do so, we perturb the embeddings and perform a new forward pass.
Finally, we combine the logits from the original and the additional forward pass, producing a final output incorporating both perspectives.

By selectively reframing inputs, \ours mimics human reconsideration under uncertainty, achieving up to 5.16\% performance improvements with minimal additional cost.
Our method outperforms beam search \citep{kasai2024acall} across multiple tasks, offering higher accuracy and significantly faster inference.
The code\footnote{\url{https://github.com/romsto/HARP}} is publicly available.

Our contributions can be summarized as follows:
\begin{itemize}
    \item We introduce a selection process based on token-level uncertainty to determine when additional computation is beneficial during inference and demonstrate its importance for improving model performance.
    \item We evaluate a novel approach to reframing inputs at inference time using dropout on the embeddings.
    \item We combine these two components to create \ours (\oursfull), a method that generally applies to any decoding algorithms, and evaluate its performance across various tasks, model sizes, and in combination with existing techniques.
\end{itemize}
\section{Related Works}
\label{sec:related_works}

\subsection{Adaptive Computation in Transformers}
Adaptive computation in Transformers (ACT) can be categorized into efficiency-focused and performance-focused categories.
Efficiency-focused ACT has been the primary focus of research, aiming at improving efficiency by reducing computation for ``easier'' inference steps (\citealp{leviathan2023fast}; \citealp{chen2023accelerating}; \citealp{elhoushi-etal-2024-layerskip}).
These approaches often involve using smaller models or skipping layers when processing less challenging tokens, thereby optimizing computational resources and resulting in inference speedups.

In contrast, performance-focused ACT, which includes our method \ours, targets the ``harder'' inference steps, prioritizing performance gains over efficiency improvements.
Our approach shares motivations with the work of \citet{goyal2024think}, who allocate more computation by extending the model's vocabulary with pause tokens.
These prepend tokens to the generation instruct the model to perform additional fixed computations, resulting in higher accuracy.
However, while effective, the pause tokens method requires retraining and fine-tuning of the model.

Unlike the pause tokens approach or scaled-up models using efficiency-focused ACT, \ours is a training-free, model-agnostic, and plug-and-play method that can be applied to any Transformer-based model without the need for retraining, making it an advantageous solution within the performance-focused ACT framework.

In parallel to adaptive computation work, some studies have explored methods to enhance reasoning capabilities in LLMs (\eg \citealp{zelikman2022star}; \citealp{zelikman2024quietstar}; \citealp{hosseini2024vstar}; \citealp{andukuri2024stargate}).
While these approaches can be seen as incorporating extra computation, their focus diverges from ours, as they improve reasoning through fine-tuning rather than optimizing token-level computation selectively.

\subsection{Uncertainty Estimation in Language Modeling}
Uncertainty quantification \citep{abdar_review_2020} on token-level is not a well-explored area.
Most of the existing works focus on the evaluation of sequence-level uncertainty (\eg \citealp{arteaga2024hallucinationdetectionllmsfast}; \citealp{manakul-etal-2023-selfcheckgpt}; \citealp{kuhn2023semantic}) or on a higher level.
In contrast, our work focuses on token-level uncertainty---the uncertainty in the probability distribution over the vocabulary for predicting the next token.
\citet{luo2024sed} introduce a ratio-based method to measure uncertainty.
While this approach offers an intuitive interpretation of uncertainty, it lacks some theoretical grounding and might fail to capture more subtle hesitation as it relies on only the two highest probabilities of the distribution.
Therefore, we use the Shannon Entropy \citep{shannon} as our uncertainty estimator.
It is an information-theoretic uncertainty measure that captures the amount of information in a probability distribution.
Entropy represents the expected number of bits required to resolve the uncertainty of a prediction.
Higher entropy indicates more uncertainty, while lower entropy suggests a more confident prediction.

\subsection{Reframing at Inference Time}
Reframing data into different perspectives is a well-established technique for training machine learning models, particularly through Multi-View Learning (MVL) \citep{chaudhuri2009multiview}.
In MVL, models are trained on multiple representations of the same data, which improves generalization.
However, these techniques are restricted to the training phase and are not designed to be applied during inference \citep{xu2013surveymultiviewlearning}, our target.

Parallely, NEFTune \citep{jain2024neftune} brings a promising direction.
It introduces noise into embeddings during the training to improve instruction-based fine-tuning.
Although NEFTune targets the training phase only, we hypothesize that a similar approach---injecting noise into embeddings---could be beneficial during inference as well.
By adding noise into embeddings at inference time, the model could gain a new perspective of the same inputs, potentially improving its ability to handle ambiguous inputs.
While NEFTune uses random uniform noise, our work explores different noise approaches, utilizing dropout on the embeddings to induce a new representation.
As detailed in Appendix \ref{adx:neftune}, we find that dropout leads to more consistent improvements.

\section{Methods}
\label{sec:methods}

In this section, we present our \oursfull (\ours) method.
We aim to perform an additional forward step from a different perspective when the model encounters uncertainty.
We begin by reviewing the standard forward pass of transformers.
Then, we introduce the two key components of \ours: quantifying uncertainty during inference and reframing inputs.
Finally, we will integrate these components to present the complete algorithm.

\subsection{Preliminary:  Transformer Forward Pass}
\label{sec:standard}

We recall the forward pass of a Transformer model as we will modify its architecture.
It processes input tokens to generate logits, representing unnormalized probabilities for predicting each token in the sequence, including the next token.
Let $\mathbf{x} = (x_1, \ldots, x_n)$ be the tokenized input sequence of length $n$, where each $x_i$ is a token ID.
For simplicity, we denote the embedding layer as $emb(\cdot)$, while $f_{\setminus emb}(\cdot)$ represents the rest of the model, consisting of $N$ serial layers.
Each layer includes a multi-head self-attention sublayer, a fully connected sublayer, and layer normalizations.
We deliberately omit the discussion of positional encodings and last-layer projection.
First, the embedding layer $emb$ maps each input token to a dense vector representation: $\mathbf{e} = emb(\mathbf{x})$, where $\mathbf{e} \in \mathbb{R}^{n \times d}$ and $d$ is the embedding dimension.
The embedded inputs are then processed through the rest of the model.

Thus, the forward pass can be concisely expressed as:
\begin{equation}
logits = f_{\setminus emb}(emb(\mathbf{x}))
\end{equation}
where $logits \in \mathbb{R}^{n \times |V|}$ and $|V|$ is the vocabulary size.
The resulting $logits$ contains unnormalized predictions for each input position.
The last position's logits are used to predict the next token.

\subsection{Uncertainty Estimation}
\label{sec:esti_unc}

We want to quantify the model's uncertainty for each newly generated token.
To do this, we focus on the logits of the last position, which are used to predict the next token.
First, the logits are normalized using the $\textproc{Softmax}$ function to obtain a probability distribution $\mathbb{P}$ over the vocabulary $V$.

The $\textproc{Softmax}$ ensures that each value in the distribution is between 0 and 1 and that the total sum of probabilities over $V$ equals 1, \ie, $\sum_{i=1}^{|V|} P(v_i\ |\ \mathbf{x}) = 1$, where $v_i$ represents a token in the vocabulary $V$ and $P(v_i\ |\ \mathbf{x})$ is the probability of token $v_i$, knowing $\mathbf{x}$, under the distribution $\mathbb{P}$.

To measure the uncertainty of the distribution $\mathbb{P}$, we then use Shannon entropy \citep{shannon}.
The entropy $\textproc{Shannon}$ is defined as:
\begin{equation}
    \textproc{Shannon}(\mathbb{P}) = -\sum_{i=1}^{|V|} P(v_i\ |\ \mathbf{x}) \log_2 P(v_i\ |\ \mathbf{x})
\end{equation}

\subsection{Reframing Inputs}
\label{sec:reframing}

Our objective is to perturb the embeddings, which represent the model's understanding of the data, to present the input sequence to the model from an alternate perspective.

We follow the approach of NEFTune \citep{jain2024neftune}, which injects random noise into the embeddings during training.
NEFTune generates this noise by sampling values from a uniform distribution in $[-1, 1]$, then scaling the noise based on sequence length, embedding dimension, and a tunable parameter.
To explore alternative perturbation methods, we experimented with various noise strategies.
We choose to use dropout among them, as it consistently yields the best results.
In Appendix \ref{adx:neftune}, we demonstrate that dropout performs better than NEFTune's random uniform noise in this context.

Let $\mathbf{e} = emb(\mathbf{x})$ be the original embeddings of the input sequence $\mathbf{x}$, and let $\delta$ be the dropout rate.
The reframed embeddings $\mathbf{\hat{e}}$ are then obtained as follows:
\begin{equation}
    \mathbf{\hat{e}} = \textproc{Dropout}(\mathbf{e}, \delta)
\end{equation}
where $\textproc{Dropout}$ randomly sets a fraction $\delta$ of the elements in the embeddings to zero.

\begin{algorithm}
\caption{\ours: \oursfullletter Step}
\label{alg:fp}
\begin{algorithmic}[1]
\Require input sequence $\mathbf{x}$, embedding layer $emb(\cdot)$, rest of the model $f_{\setminus emb}(\cdot)$
\Ensure uncertainty threshold $\theta$, drop-out rate $\delta$, combination factor $\beta$
\State $\mathbf{e} \gets emb(\mathbf{x})$ 
\State $logits \gets f_{\setminus emb}(\mathbf{e})$ 
\State $\mathbb{P} \gets \textproc{Softmax}(logits)$ 
\LineComment{If the model is uncertain, we perform one more forward pass but reframed.}
\If{$\textproc{Shannon}(\mathbb{P}) > \theta$}
    \State $\mathbf{\hat{e}} \gets \textproc{DropOut}(\mathbf{e}, \ \delta)$ 
    \State $logits_r \gets f_{\setminus emb}(\mathbf{\hat{e}})$ 
    \State $logits \gets \beta * logits + (1 - \beta) * logits_r$ 
\EndIf
\State \Return $logits$
\end{algorithmic}
\end{algorithm}
\subsection{\oursfull (\ours)}

Our method, \ours, adapts the standard Transformer forward pass by adding an additional computation step when the model exhibits uncertainty.
This procedure is outlined in Algorithm \ref{alg:fp}, using the same colors as in Figure \ref{fig:bigpicture} for clarity.

First, we perform the standard Transformer forward pass (presented in Section \ref{sec:standard}).
From the input sequence $\mathbf{x}$, we compute the token embeddings $\mathbf{e}$ and the logits, denoted as $logits$.

Next, we estimate the uncertainty at the current step using Shannon Entropy (detailed in Section \ref{sec:esti_unc}).
To achieve this, we normalize the logits and compute the Shannon entropy.

If the entropy is below a predefined uncertainty threshold $\theta$, the model is considered confident in its generation.
In this case, we return the logits from the standard forward pass, and the current generation step ends.

However, if the uncertainty exceeds the threshold $\theta$, the model is uncertain.
In this case, we initiate a second forward pass but with a reframed input sequence (explained in Section \ref{sec:reframing}).
The reframed embeddings, $\mathbf{\hat{e}}$, are obtained by applying dropout with a rate $\delta$ to the original embeddings $\mathbf{e}$.
We then perform a second forward pass using $\mathbf{\hat{e}}$, which outputs reframed logits, denoted as $logits_r$.
This additional forward pass requires recomputation as it cannot leverage the KVCache (see Section \ref{sec:limitations}).

Finally, the original and reframed logits are combined using a combination factor $\beta$ to produce the final output logits.
\begin{equation}
    logits = \beta \cdot logits + (1 - \beta) \cdot logits_r
    \label{eq:combine}
\end{equation}
The combination factor $\beta$ controls the balance between original and reframed logits.
We empirically find that setting $\beta = 0.5$ balances the contributions of both passes effectively.

The dropout-based perturbation forces the model to consider different representations of the input.
By selectively introducing additional computations only when necessary, \ours balances computational efficiency and prediction reliability.

Our method introduces three hyperparameters: the uncertainty threshold $\theta$, the dropout rate $\delta$, and the combination factor $\beta$.
In our experiments, we set these parameters to values identified as optimal during preliminary testing.
The optimal choices for these hyperparameters may vary across tasks or models and can benefit from further tuning;
we provide a preliminary study on the impact of $\theta$ in Appendix \ref{adx:theta_adx} and leave a more in-depth exploration of all hyperparameters for future work.

\section{Experimental Set-Up}
\label{sec:experiments}

\subsection{Models}
We consider decoder-only models of size 3.8B, 7B, and 8B as they are the standard in recent times.
Particularly, we use LLaMA-3.1 \citep{dubey2024llama3herdmodels}, Mistral 7B v0.3 \citep{jiang2023mistral7b} and Phi 3.5 Mini \citep{abdin2024phi3technicalreporthighly} to cover different scales and architectures.
We consider only aligned models (denoted as "\emph{Instruct}") for their simplicity of evaluation and greater performance.
\emph{Aligned models} \citep{christiano2017deep} refer to models that are fine-tuned to better follow instructions, using methods such as supervised fine-tuning or reinforcement learning from human feedback (RLHF).
All the models are loaded using quantized INT8 precision to fit the GPU and accelerate inference.

\subsection{Datasets}
As our method targets ``off-the-shelf'' LLMs, we consider five datasets covering varied downstream tasks and output formats in order to reproduce the variety of missions they can encounter.

\begin{itemize}
    \item \textbf{GSM8K} \citep{cobbe2021training} is a famous mathematical reasoning benchmark where the task involves solving grade school-level math word problems.
    The output is a free text, typically a detailed solution or numerical answer.
    \item \textbf{CommonSenseQA (CsQA)} \citep{talmor-etal-2019-commonsenseqa} is a multiple-choice question benchmark that tests commonsense reasoning and general understanding abilities of the model.
    The output is one selected option from five possible choices.
    \item \textbf{LAMBADA} \citep{paperno-etal-2016-lambada}, a language modeling benchmark focused on text understanding and completion, where the task is to predict the final word of a passage based on its context.
    The output is a single word.
    \item \textbf{MMLU Pro} \citep{wang2024mmlupro}---an enhancement of Massive Multitask Language Understanding (MMLU) \citep{hendrycks2021measuring} dataset---evaluates models on a wide range of academic and professional subjects.
    The task consists of answering multiple-choice questions, with the output being one selected option from up to ten possible choices.
    \item \textbf{CNN/Daily Mail (CNN/DM)} \citep{nallapati-etal-2016-abstractive-nobug} dataset is used for text summarization tasks, where the goal is to generate concise summaries of news articles. The output is free text in the form of a summary.
\end{itemize}

Prompt details can be found in Appendix \ref{adx:prompts}.
All the datasets are evaluated with a zero-shot prompt.

\subsection{Evaluation}
Since \ours is generally applicable to any decoding methods, we compare different decoding methods \citep{shi2024thorough} both with and without the \ours forward pass modification.
We use greedy search decoding---a deterministic method that fetches the highest probable token---and nucleus sampling search---a stochastic one that samples the next token \citep{shi2024thorough}.
In addition, we evaluate beam search decoding \citep{kasai2024acall} without \ours as this decoding method usually results in more accurate results.
\begin{table*}[t]
\centering
\scalebox{0.82}{
\begin{tabular}{l|c|ccccc|c}
\toprule
\multicolumn{1}{c|}{\multirow{2}{*}{\textbf{Models}}} & \multicolumn{1}{c|}{\multirow{2}{*}{\textbf{Methods}}} & \multicolumn{5}{c|}{\textbf{Datasets}} & \multicolumn{1}{c}{\multirow{2}{*}{\shortstack{\textbf{Relative Cost} \\ {\small overhead} \\ {\small to Vanilla}}}} \\
 & & CsQA & GSM8K & LAMBADA & MMLU Pro & CNN/DM & \\ \midrule
\multirow{5}{*}{\shortstack{\textbf{LLaMA-3.1} \\ {\small Instruct (8B)}}} & Vanilla (Greedy) & 78.79 & \underline{76.88} & 30.86 & 46.42 & 32.44 & \\
 & Beam Search & \underline{79.29} & 76.38 & \underline{31.35} & \underline{48.21} & \underline{33.17} & x2.79\\
 & \cellcolor{lightgray!30}Ours (Greedy) & \textbf{80.30} \scriptsize{\textcolor{HarderGreen}{(+1.52)}} & \textbf{78.39} \scriptsize{\textcolor{HarderGreen}{(+1.51)}} & \textbf{36.02} \scriptsize{\textcolor{HarderGreen}{(+5.16)}} & \textbf{48.21} \scriptsize{\textcolor{HarderGreen}{(+1.79)}} & \textbf{34.03} \scriptsize{\textcolor{HarderGreen}{(+1.59)}} & \textbf{x1.16}\\\dashline{2}{8}
 & Vanilla (Nucleus) & \textbf{79.80} & 73.00 & 27.44 & 42.55 & 30.81 & \\
 & \cellcolor{lightgray!30}Ours (Nucleus) & 79.29 \scriptsize{\textcolor{HarderRed}{(-0.51)}} & \textbf{74.00} \scriptsize{\textcolor{HarderGreen}{(+1.00)}} & \textbf{31.38} \scriptsize{\textcolor{HarderGreen}{(+3.94)}} & \textbf{43.45} \scriptsize{\textcolor{HarderGreen}{(+0.90)}} & \textbf{32.38} \scriptsize{\textcolor{HarderGreen}{(+1.57)}} & \textbf{x1.17}\\ \midrule
\multirow{5}{*}{\shortstack{\textbf{Mistral v0.3} \\ {\small Instruct (7.25B)}}} & Vanilla (Greedy) & 70.37 & 43.62 & 45.15 & \underline{31.76} & \underline{29.11} & \\
 & Beam Search & \textbf{70.99} & \textbf{50.53} & \underline{45.70} & \textbf{33.11} & 28.71 & x3.07\\
 & \cellcolor{lightgray!30}Ours (Greedy) & \textbf{70.99} \scriptsize{\textcolor{HarderGreen}{(+0.62)}} & \underline{48.40} \scriptsize{\textcolor{HarderGreen}{(+4.79)}} & \textbf{49.76} \scriptsize{\textcolor{HarderGreen}{(+4.64)}} & \underline{31.76} \scriptsize{(0.00)} & \textbf{29.57} \scriptsize{\textcolor{HarderGreen}{(+0.47)}} & \textbf{x1.24}\\\dashline{2}{8}
 & Vanilla (Nucleus) & \textbf{70.74} & 29.38 & 45.02 & 31.76 & 28.72 & \\
 & \cellcolor{lightgray!30}Ours (Nucleus) & 70.21 \scriptsize{\textcolor{HarderRed}{(-0.53)}} & \textbf{31.88} \scriptsize{\textcolor{HarderGreen}{(+2.50)}} & \textbf{48.26} \scriptsize{\textcolor{HarderGreen}{(+3.24)}} & \textbf{33.79} \scriptsize{\textcolor{HarderGreen}{(+2.03)}} & \textbf{28.72} \scriptsize{(0.00)} & \textbf{x1.29}\\\midrule
\multirow{5}{*}{\shortstack{\textbf{Phi 3.5 Mini} \\ {\small Instruct (3.82B)}}} & Vanilla (Greedy) & 77.20 & 72.50 & 32.76 & 29.65 & \underline{26.10} & \\
 & Beam Search & \underline{77.72} & \underline{73.00} & \textbf{33.60} & \textbf{33.78} & 25.79 & x3.18\\
 & \cellcolor{lightgray!30}Ours (Greedy) & \textbf{78.24} \scriptsize{\textcolor{HarderGreen}{(+1.04)}} & \textbf{73.00} \scriptsize{\textcolor{HarderGreen}{(+0.50)}} & \underline{33.44} \scriptsize{\textcolor{HarderGreen}{(+0.68)}} & \underline{32.75} \scriptsize{\textcolor{HarderGreen}{(+3.10)}} & \textbf{26.97} \scriptsize{\textcolor{HarderGreen}{(+0.87)}} & \textbf{x1.26}\\\dashline{2}{8}
 & Vanilla (Nucleus) & 77.04 & 71.50 & 31.85 & 28.82 & \textbf{25.30} & \\
 & \cellcolor{lightgray!30}Ours (Nucleus) & \textbf{77.55} \scriptsize{\textcolor{HarderGreen}{(+0.51)}} & \textbf{74.50} \scriptsize{\textcolor{HarderGreen}{(+3.00)}} & \textbf{32.99} \scriptsize{\textcolor{HarderGreen}{(+1.14)}} & \textbf{34.53} \scriptsize{\textcolor{HarderGreen}{(+5.71)}} & 25.19 \scriptsize{\textcolor{HarderRed}{(-0.11)}} & \textbf{x1.27}\\
\bottomrule
\end{tabular}
}
\caption{Performance comparison of the original model (Vanilla) with various decoding methods: greedy search, nucleus sampling, beam search, and our \ours modified forward pass in both greedy and nucleus sampling settings. Numbers in parentheses indicate performance \textcolor{HarderGreen}{gain} or \textcolor{HarderRed}{loss} relative to Vanilla for each decoding method. The cost column shows the relative inference time based on Vanilla's corresponding decoding method, averaged over five datasets. Reported scores reflect accuracy, except for CNN/DM, where the ROUGE-1 score is reported.}
\label{tab:main}
\end{table*}
The hyperparameters used in our experiments are fixed at a dropout rate of $\delta = 0.20$ and an uncertainty threshold of $\theta = 1.0$.
We set the temperature for nucleus sampling to $\tau = 0.6$ and the top-$p$ to 0.9.
In the case of beam search, the number of beams is $b = 3$ with a top-$k$ of 5.
We chose to apply length-normalization \citep{wu2016google} as we found it beneficial to tasks such as summarization and reasoning.
This length penalty is fixed to $\alpha = 0.6$ for every datasets, as we seek to compare \ours with model- and task-agnostic methods.
We empirically found the value by evaluating a model on a subset of three datasets (CNN/DM, GSM8K and LAMBADA).
When evaluating Chain-of-Thought \citep{wei2024cot} prompting, we prepend ``\texttt{Let's think step-by-step.}'' to the generation of the model.
We evaluate a subset of each dataset with a batch size of 1, without caching (KVCache), on a single RTX3090 (24GB) using the same seed for every example.
The accuracy is reported for all datasets except CNN/DM, where we evaluate the ROUGE-1 score.
Moreover, we record the generation time of each method.
\section{Results}
\label{sec:results}


\paragraph{\ours improves the performance of all tasks.}
Our method demonstrates consistent performance improvements across all tasks.
As shown in Table \ref{tab:main}, in the greedy setting, \ours outperforms both the vanilla model and beam search decoding in most scenarios.
The improvements are particularly notable for tasks like LAMBADA, with gains of up to 5.16\%.
When applied to nucleus sampling, \ours outperforms the vanilla model in most cases, especially in math-reasoning tasks like GSM8K.

On GSM8K, \citet{goyal2024think} achieved a +1.00\% improvement (from 7.50\% to 8.50\% accuracy) with their training and fine-tuning pause token method.
In comparison, \ours delivers an even higher improvement of +1.51\% on the same dataset, using a more performant model and without any retraining.

In summary, our \ours allows further improvements as high as +5.16\% of high-end models that already show strong performance on many tasks without requiring any fine-tuning.
This demonstrates the potential of \ours to enhance state-of-the-art models.

\paragraph{\ours is model-agnostic.}
We show that \ours also consistently improves models ranging from 3 to 8 billion parameters, including  LLaMA-3.1, Mistral v0.3, and Phi 3.5 Mini.
It illustrates its robustness and wide applicability across diverse language models.


\begin{table}[ht]
\centering
\vspace{2mm}
\scalebox{0.75}{
\begin{tabular}{l|c|ccc}
\toprule
\multicolumn{1}{c|}{\multirow{2}{*}{\textbf{Models}}} & \multicolumn{1}{c|}{\multirow{2}{*}{\textbf{Methods}}} & \multicolumn{3}{c}{\textbf{Datasets}} \\
 & & CsQA & GSM8K & MMLU Pro \\ \midrule
 \multirow{2}{*}{\textbf{LLaMA}} & CoT & 74.95 & 75.48 & 48.19\\
 & \cellcolor{lightgray!30}Ours CoT & \textbf{75.75} \scriptsize{\textcolor{HarderGreen}{(+0.80)}} & \textbf{76.58} \scriptsize{\textcolor{HarderGreen}{(+1.10)}} & \textbf{49.35} \scriptsize{\textcolor{HarderGreen}{(+1.16)}}\\
 \midrule
\multirow{2}{*}{\textbf{Mistral}} & CoT & 70.64 & \textbf{48.00} & 32.37\\
 & \cellcolor{lightgray!30}Ours CoT & \textbf{75.23} \scriptsize{\textcolor{HarderGreen}{(+4.59)}} & 46.00 \scriptsize{\textcolor{HarderRed}{(-2.00)}} & \textbf{32.95} \scriptsize{\textcolor{HarderGreen}{(+0.58)}}\\
\bottomrule
\end{tabular}
}
\caption{Accuracy comparison of Chain-of-Thought (CoT) and \ours applied to CoT, using greedy decoding. Numbers in parentheses indicate \textcolor{HarderGreen}{gain} or \textcolor{HarderRed}{loss} relative to the standard CoT approach.}
\label{tab:cot}
\vspace{-2mm}
\end{table}

\paragraph{\ours works with advanced techniques.}
As shown in Table \ref{tab:cot}, advanced techniques, such as Chain-of-Thought (CoT) \citep{wei2024cot} prompting, are further enhanced with \ours.
For instance, applying the modified forward pass to CoT prompting with the LLaMA model improves accuracy by 0.8\%, 1.16\%, and 1.10\% on CommonsenseQA, MMLU Pro, and GSM8K, respectively.
This shows our method can seamlessly combine with other existing methods, further enhancing performance.


\begin{figure}[t]
    \centering
    \includegraphics[width=\linewidth]{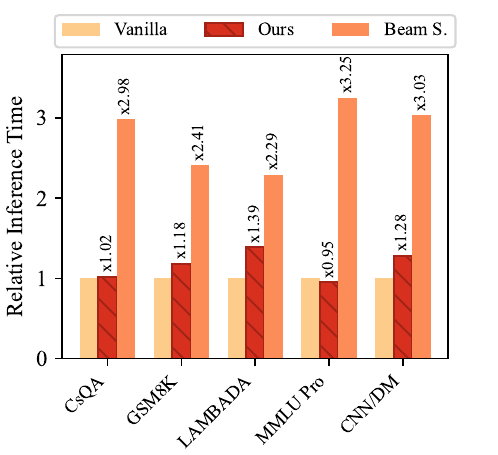}
    \caption{LLaMA 3.1 Instruct (8B) average relative inference time of the original model greedy search (Vanilla), beam search decoding (Beam S.), and our \ours (Ours) using greedy search decoding. Values and other models are detailed in Table \ref{tab:inf_speed}.}
    \label{fig:inference_speed}
    \vspace{-2mm}
\end{figure}

\paragraph{\ours is faster than Beam Search.}
Figure \ref{fig:inference_speed} reveals that despite the additional computation, the inference time of \ours remains close to that of the vanilla generation, introducing a minimal additional computational cost.
The inference time for our method is consistently under twice that of the original model without caching across all tasks.
In contrast, beam search---while most of the time yielding lower performance than \ours---induces a significantly higher inference cost.
This is the indicator that our method strikes a balance between performance gains and inference time.

Beam search shows significant delays, often higher than x2.4 the time required by the greedy search generation, whereas \ours achieves inference times lower than x1.4.
Additional results in Table \ref{tab:inf_speed} validate this efficiency by comparing other models.
For example, the Mistral model increases inference time by only x1.17 to x1.49, while beam search increases it by x2.51 to x3.72, compared to the original model without KVCache.

On average, our method results in an x1.25 increase in inference time across all tasks and models, demonstrating its efficiency compared to more computationally intensive approaches like beam search, which takes more than twice as long.
\section{Analysis}
\label{sec:analysis}

\begin{table*}[ht]
\centering
\scalebox{0.84}{
\begin{tabular}{l|c|ccccc|c}
\toprule
\multicolumn{1}{c|}{\multirow{2}{*}{\textbf{Models}}} & \multicolumn{1}{c|}{\multirow{2}{*}{\textbf{Methods}}} & \multicolumn{5}{c|}{\textbf{Datasets}} & \multicolumn{1}{c}{\multirow{2}{*}{\textbf{Relative Cost}}}\\
 & & CsQA & GSM8K & LAMBADA & MMLU Pro & CNN/DM & \\ \midrule
\multirow{2}{*}{\shortstack{\textbf{LLaMA-3.1} \\ {\small Instruct (8B)}}} & w\textbackslash o \textproc{Shannon} & 77.17 \scriptsize{\textcolor{HarderRed}{(-1.62)}} & \textbf{78.39} \scriptsize{\textcolor{HarderGreen}{(+1.51)}} & \textbf{36.08} \scriptsize{\textcolor{HarderGreen}{(+5.22)}} & 44.88 \scriptsize{\textcolor{HarderRed}{(-1.50)}} & 32.66 \scriptsize{\textcolor{HarderGreen}{(+0.22)}} & x2.31\\
 & w\textbackslash\ \textproc{Shannon} & \textbf{80.30} \scriptsize{\textcolor{HarderGreen}{(+1.52)}} & \textbf{78.39} \scriptsize{\textcolor{HarderGreen}{(+1.51)}} & 36.02 \scriptsize{\textcolor{HarderGreen}{(+5.16)}} & \textbf{48.21} \scriptsize{\textcolor{HarderGreen}{(+1.79)}} & \textbf{34.03} \scriptsize{\textcolor{HarderGreen}{(+1.59)}} & \textbf{x1}\\ \midrule
 
\multirow{2}{*}{\shortstack{\textbf{Mistral v0.3} \\ {\small Instruct (7.25B)}}} & w\textbackslash o \textproc{Shannon} & 70.37 \scriptsize{(0.00)} & 45.74 \scriptsize{\textcolor{HarderGreen}{(+2.12)}} & 49.56 \scriptsize{\textcolor{HarderGreen}{(+4.41)}} & \textbf{31.76} \scriptsize{(0.00)} & 28.62 \scriptsize{\textcolor{HarderRed}{(-0.49)}} & x1.68\\
 & w\textbackslash\ \textproc{Shannon} & \textbf{70.99} \scriptsize{\textcolor{HarderGreen}{(+0.62)}} & \textbf{48.40} \scriptsize{\textcolor{HarderGreen}{(+4.79)}} & \textbf{49.76} \scriptsize{\textcolor{HarderGreen}{(+4.64)}} & \textbf{31.76} \scriptsize{(0.00)} & \textbf{29.57} \scriptsize{\textcolor{HarderGreen}{(+0.47)}} & \textbf{x1}\\
\bottomrule
\end{tabular}
}

\caption{Study of the impact of unconditional additional step (w\textbackslash o \textproc{Shannon}) and \ours uncertainty-based additional step (w\textbackslash\ \textproc{Shannon}) using greedy decoding. The cost column denotes the relative inference time with w\textbackslash\ \textproc{Shannon} averaged over the five datasets. Numbers in parentheses indicate performance \textcolor{HarderGreen}{gain} or \textcolor{HarderRed}{loss} relative to the original model performance.}
\label{tab:th0}
\vspace{-2mm}
\end{table*}
\begin{table*}[ht]
\centering
\scalebox{0.84}{
\begin{tabular}{l|c|ccccc|c}
\toprule
\multicolumn{1}{c|}{\multirow{2}{*}{\textbf{Models}}} & \multicolumn{1}{c|}{\multirow{2}{*}{\textbf{Methods}}} & \multicolumn{5}{c|}{\textbf{Datasets}} & \multicolumn{1}{c}{\multirow{2}{*}{\textbf{Relative Cost}}}\\
 & & CsQA & GSM8K & LAMBADA & MMLU Pro & CNN/DM & \\ \midrule
\multirow{3}{*}{\shortstack{\textbf{LLaMA-3.1} \\ {\small Instruct (8B)}}} & 1-step & \textbf{80.30} & 78.39 & \textbf{36.02} & \textbf{48.21} & 34.03 & \textbf{x1}\\
 & 2-steps & 79.80 \scriptsize{\textcolor{HarderRed}{(-0.50)}} & 76.88 \scriptsize{\textcolor{HarderRed}{(-1.49)}} & 35.52 \scriptsize{\textcolor{HarderRed}{(-0.50)}} & 47.19 \scriptsize{\textcolor{HarderRed}{(-1.02)}} & 33.92 \scriptsize{\textcolor{HarderRed}{(-0.11)}} & x1.52\\
 & 4-steps & 79.70 \scriptsize{\textcolor{HarderRed}{(-0.60)}} & \textbf{80.40} \scriptsize{\textcolor{HarderGreen}{(+2.01)}} & 35.02 \scriptsize{\textcolor{HarderRed}{(-1.00)}} & 43.73 \scriptsize{\textcolor{HarderRed}{(-4.48)}} & \textbf{34.29} \scriptsize{\textcolor{HarderGreen}{(+0.26)}} & x1.97\\
   \midrule
 
\multirow{3}{*}{\shortstack{\textbf{Mistral v0.3} \\ {\small Instruct (7.25B)}}} & 1-step & \textbf{70.99} & \textbf{48.40} & \textbf{49.76} & \textbf{31.76} & 29.57 & \textbf{x1}\\
 & 2-steps & 70.47 \scriptsize{\textcolor{HarderRed}{(-0.52)}} & 46.38 \scriptsize{\textcolor{HarderRed}{(-2.02)}} & 48.26 \scriptsize{\textcolor{HarderRed}{(-1.50)}} & \textbf{31.76} \scriptsize{(0.00)} & 29.66 \scriptsize{\textcolor{HarderGreen}{(+0.09)}} & x1.11\\
 & 4-steps & 70.37 \scriptsize{\textcolor{HarderRed}{(-0.62)}} & 45.32 \scriptsize{\textcolor{HarderRed}{(-3.08)}} & 48.15 \scriptsize{\textcolor{HarderRed}{(-1.61)}} & \textbf{31.76} \scriptsize{(0.00)} & \textbf{29.83} \scriptsize{\textcolor{HarderGreen}{(+0.26)}} & x1.37\\
\bottomrule
\end{tabular}
}
\caption{Performance comparison using a different number of reframing steps across all the datasets. "$x$-steps" indicates that the model can perform up to $x$ additional forward passes when uncertainty exceeds the threshold. The cost column denotes the average relative inference time with \ours (1-step). Numbers in parentheses indicate performance \textcolor{HarderGreen}{gain} or \textcolor{HarderRed}{loss} relative to \ours.}

\label{tab:msteps}
\vspace{-2mm}
\end{table*}
\paragraph{Uncertainty guides the additional computations.}

To investigate the effectiveness of our uncertainty-based approach (denoted as `w\textbackslash\ \textproc{Shannon}'), we compare it with a method that unconditionally adds an extra forward step for every token (denoted as `w\textbackslash o \textproc{Shannon}'). 
This comparison helps us understand whether the improvements come solely from the additional computation or from our uncertainty-selected approach.
Table \ref{tab:th0} presents the results across multiple tasks and two models.

Our analysis reveals that unconditionally adding an extra forward step is not universally beneficial.
While it improves performance compared to the vanilla model in most cases, especially in tasks like GSM8K and LAMBADA, it also results in lower or equal performance on tasks like multiple-choice questions (CommonsenseQA and MMLU Pro).

The uncertainty-conditional method outperforms the unconditional approach, even though it is sometimes marginal.
This suggests that uncertainty is an excellent signal for selecting the steps requiring extra computation.
A key advantage of the selective approach is its computational efficiency.
As shown in the Cost column of Table \ref{tab:th0}, the uncertainty-based method requires significantly lower computational overhead than the unconditional method. 

While additional computation can generally improve performance, our uncertainty-based approach offers a more nuanced and efficient solution.
It achieves comparable or superior results to unconditional computation while balancing performance gains and computational costs.

\paragraph{Uncertainty pinpoints key reasoning steps.}
We analyze the output generations of \ours by highlighting the tokens that require extra computation in Appendix \ref{adx:uncloc}.
Upon reviewing some examples, particularly in problem-solving tasks, we observe that high-uncertainty states usually appear at the start of each reasoning step.
This mirrors human decision-making, where individuals take more time to consider the stages of reasoning to construct a valid response than the actual content of the reasoning.

Additionally, we note that \ours tends to generate shorter sequences than the original forward pass.
Across every task (except ones with next-word prediction), we observe an average reduction of 5.5\% in output length when applying \ours.
This shortening effect is not caused by promoting the end-of-sequence ($EOS$) token as the most likely token earlier in uncertain cases.
Instead, this effect appears from variations in token selection during earlier stages of generation.
\ours results in more concise responses, especially for reasoning-based tasks and summarizing tasks.

\paragraph{One additional representation is sufficient for reframing.}
\label{sec:multi_steps}
When facing uncertainty, \ours reframes the inputs only once before pursuing the next generation step.
To explore the effect of multiple reframings, we experiment with adding extra forward passes while the uncertainty remains higher than $\theta$ (capping it at a maximum number of steps).

Surprisingly, Table \ref{tab:msteps} shows that increasing the number of additional steps often leads to a decline in performance.
This interesting outcome suggests that while a single additional representation can provide a useful alternative perspective on the inputs, too many representations may be penalizing.
In datasets like CommonsenseQA, LAMBADA, and MMLU Pro, we observe a consistent drop in accuracy with more reframing steps.

Although GSM8K achieves a notable +2.01\% increase in accuracy using four additional steps with LLaMA, these gains come at the cost of significantly higher computation time, scaling up to x1.97 in the 4-step method.

We hypothesize that more than two representations might confuse and distract the model from the original task, decreasing precision.
This aligns with cognitive theories, suggesting that while considering reframings enhances problem-solving, too many representations increase cognitive load \citep{sweller1988cognitive}, reducing performance.
In a more mathematical way, this behavior may be due to the random noise introduced during reframing and the way logits are combined.
While random noise can sometimes be beneficial, it can also be detrimental, and increasing the number of reframings increases the likelihood of harmful noise.
Additionally, as the logits from the vanilla and reframed representations are averaged (see Equation \ref{eq:combine}), introducing more than one reframing step reduces the weight of the vanilla logits, disproportionately favoring the reframed ones, contributing to the higher likelihood of harmful representations.
\section{Conclusion}
\label{sec:conclusion}

This paper presented a novel method, \ours, designed to enhance language model inference without requiring retraining or architectural modifications.
Our approach uses a selection process based on token-level uncertainty to identify when additional computation is advantageous and an innovative method for reframing inputs during inference.
We demonstrated that \ours can achieve significant performance improvements across various tasks and model sizes, with accuracy gains of up to 5.16\%.
Importantly, \ours maintains inference efficiency compared to methods like beam search, with only an x1.25 average increase in inference time over the standard model.
While \ours provides a promising proof of concept, its real-world application is currently limited by challenges such as cache invalidity and the introduced randomness.

\section{Limitations}
\label{sec:limitations}
Our study has several limitations that need to be addressed in future research.

First, due to resource and time constraints, our experiments were limited in scope.
Evaluations were conducted on quantized models and subsets of each dataset, and we implemented custom generation methods.
Additionally, the method has yet to be tested on larger language models, particularly those with 70 billion parameters or more.
Although we attempt to cover a range of tasks, there are other language challenges that our method has not been tested on.
Furthermore, our experiments did not leverage widely used libraries, such as LM-Evaluation-Harness \citep{eval-harness}, which could provide more standardized benchmarking.
Consequently, while initial results demonstrate potential, further work is required to confirm whether \ours generalizes effectively across tasks and scales.
To partially address this limitation, we provide an extended evaluation using LM-Evaluation-Harness in Appendix~\ref{adx:extended}.
\ours serves as a proof of concept, demonstrating the potential of uncertainty-aware adaptive computation in improving inference performance.

Furthermore, our current implementation faces some challenges in inference efficiency.
The method could slow down batch processing since uncertain tokens require additional computation.
Moreover, performance might be impacted when using \ours with models that employ key-value caching (KVCache).
Embedding dropout may temporarily invalidate the KVCache, causing a VRAM spike during each reframing step.
In the worst-case scenario---where all tokens are affected---VRAM usage could briefly double.
Other approaches could be explored to mitigate this, such as directly perturbing the KVCache to perform reframing or doing layer-specific perturbations.

Looking forward, there are promising ideas for future work, in addition to exploring alternative uncertainty measures or perturbation methods.
One intriguing direction would be to explore the application of the uncertainty selection mechanism during the fine-tuning process rather than only at inference time.
This could involve integrating a ``pause token,'' as proposed by \citet{goyal2024think}, during training to teach the model when to allocate additional computation.
Such an approach could enable models to increase computation when needed autonomously.
Another direction could involve adapting beam search decoding with hesitation and reframing.
Creating new beams when uncertainty is encountered and keeping track of all the branches may enhance the results even more.
Finally, combining the proposed method with speculative decoding \citep{leviathan2023fast, chen2023accelerating} could offer further optimizations, enabling gains in both efficiency and performance.

\section{Ethics Statement}
\label{sec:ethics}

In this work, we propose modifying the forward pass for improved performance.
While we evaluate different models on datasets, we acknowledge that we have not evaluated the impact of our method on safety, including concerns such as toxicity, bias, or content moderation.
Our goal is to enhance accuracy, but broader implications---such as generating harmful content or sensitive applications---remain unexplored.
\section*{Acknowledgements}
\label{sec:acknowledgements}

We are grateful to Jihyuk Kim, Jongho Kim, and Jongyoon Kim for their relecture, comments, and feedback.

\bibliography{custom}

\appendix
\onecolumn

\section{NEFTune versus Dropout}
\label{adx:neftune}

\begin{table*}[ht]
\centering
\scalebox{0.9}{
\begin{tabular}{c|ccccc}
\toprule
\multicolumn{1}{c|}{\multirow{2}{*}{\textbf{Methods}}} & \multicolumn{5}{c}{\textbf{Datasets}}\\
 & CsQA & GSM8K & LAMBADA & MMLU Pro & CNN/DM\\ \midrule
\ours (NEFTune$_{\alpha = 5}$) & \textbf{80.30} & \textbf{78.89} & 32.69 & 66.26 & 43.03\\
\ours (NEFTune$_{\alpha = 10}$) & \textbf{80.30} & 78.39 & \textbf{38.23} & 66.26 & 43.03\\
\ours (\textproc{Dropout}) & \textbf{80.30} & 78.39 & 36.02 & \textbf{71.43} & \textbf{48.21}\\
\bottomrule
\end{tabular}
}
\caption{Comparison of NEFTune noise (with hyperparameter $\alpha$ in $(5, 10)$) and dropout ($\delta = 20\%$) in the \ours method using LLaMA-3.1 Instruct 8B with greedy decoding. We report accuracy for each dataset, except CNN/DM, for which we report the ROUGE-1 score.}
\label{tab:neftune}
\end{table*}

Table \ref{tab:neftune} compares \ours using NEFTune noise and dropout.
It reveals that while NEFTune improves performance compared to the vanilla model (Table \ref{tab:main}) by adding scaled uniform noise to embeddings, dropout offers more consistent enhancements.
Specifically, dropout performances better on MMLU Pro and CNN/DM while having a close score to NEFTune on other datasets.
Moreover, NEFTune$_{\alpha=5}$ results in a lower score than the vanilla forward pass in the LAMBADA next word prediction task.
Dropout is a better candidate for infusing a different perspective representation into the model during inference time.
\begin{table*}[t]
\centering
\scalebox{0.88}{

\begin{tabular}{l|c|ccccc}
\toprule
\multicolumn{1}{c|}{\multirow{2}{*}{\textbf{Models}}} & \multicolumn{1}{c|}{\multirow{2}{*}{\textbf{Methods}}} & \multicolumn{5}{c}{\textbf{Datasets}} \\
 & & CsQA & GSM8K & LAMBADA & MMLU Pro & CNN/DM \\ \midrule
\multirow{7}{*}{\shortstack{\textbf{LLaMA 3.1} \\ {\small Instruct (8B)}}} 
& Vanilla (Greedy) & \textbf{1.09} & \textbf{47.01} & \textbf{0.80} & 24.46 & \textbf{128.39}\\
& Beam Search & 3.24 \scriptsize{\textcolor{HarderRed}{(x2.98)}} & 113.27 \scriptsize{\textcolor{HarderRed}{(x2.41)}} & 1.84 \scriptsize{\textcolor{HarderRed}{(x2.29)}} & 79.59 \scriptsize{\textcolor{HarderRed}{(x3.25)}} & 388.72 \scriptsize{\textcolor{HarderRed}{(x3.03)}}\\
& \cellcolor{lightgray!30}Ours (Greedy) & \underline{1.11} \scriptsize{\textcolor{HarderGreen}{(x1.02)}} & \underline{55.28} \scriptsize{\textcolor{HarderOrange}{(x1.18)}} & \underline{1.11} \scriptsize{\textcolor{HarderOrange}{(x1.39)}} & \underline{23.20} \scriptsize{\textcolor{HarderGreen}{(x0.95)}} & \underline{164.37} \scriptsize{\textcolor{HarderOrange}{(x1.28)}}\\ \dashline{2}{7}
& Vanilla (Nucleus) & \textbf{1.09} & \textbf{47.28} & \textbf{0.76} & \textbf{28.13} & \textbf{125.53}\\
& \cellcolor{lightgray!30}Ours (Nucleus) & 1.12 \scriptsize{\textcolor{HarderGreen}{(x1.02)}} & 55.04 \scriptsize{\textcolor{HarderOrange}{(x1.16)}} & 1.11 \scriptsize{\textcolor{HarderOrange}{(x1.46)}} & 24.36 \scriptsize{\textcolor{HarderGreen}{(x0.87)}} & 169.19 \scriptsize{\textcolor{HarderOrange}{(x1.35)}}\\ \dashline{2}{7}
& Vanilla (CoT) & \textbf{22.28} & \textbf{41.92} & - & \textbf{49.75} & -\\
& \cellcolor{lightgray!30}Ours (CoT) & 31.97 \scriptsize{\textcolor{HarderOrange}{(x1.43)}} & 49.99 \scriptsize{\textcolor{HarderOrange}{(x1.19)}} & - & 63.98 \scriptsize{\textcolor{HarderOrange}{(x1.29)}} & -\\ \midrule

\multirow{7}{*}{\shortstack{\textbf{Mistral v0.3} \\ {\small Instruct (7.25B)}}} 
& Vanilla (Greedy) & \textbf{13.63} & \textbf{95.85} & \textbf{1.40} & \textbf{112.00} & \textbf{187.57}\\
& Beam Search & 50.72 \scriptsize{\textcolor{HarderRed}{(x3.72)}} & 240.47 \scriptsize{\textcolor{HarderRed}{(x2.51)}} & 4.40 \scriptsize{\textcolor{HarderRed}{(x3.16)}} & 319.76 \scriptsize{\textcolor{HarderRed}{(x2.86)}} & 578.84 \scriptsize{\textcolor{HarderRed}{(x3.09)}}\\
& \cellcolor{lightgray!30}Ours (Greedy) & \underline{16.16} \scriptsize{\textcolor{HarderOrange}{(x1.19)}} & \underline{111.94} \scriptsize{\textcolor{HarderOrange}{(x1.17)}} & \underline{1.97} \scriptsize{\textcolor{HarderOrange}{(x1.41)}} & \underline{133.16} \scriptsize{\textcolor{HarderOrange}{(x1.19)}} & \underline{231.45} \scriptsize{\textcolor{HarderOrange}{(x1.23)}}\\ \dashline{2}{7}
& Vanilla (Nucleus) & \textbf{16.19} & \textbf{99.08} & \textbf{1.36} & \textbf{110.82} & \textbf{198.16}\\
& \cellcolor{lightgray!30}Ours (Nucleus) & 20.77 \scriptsize{\textcolor{HarderOrange}{(x1.28)}} & 121.23 \scriptsize{\textcolor{HarderOrange}{(x1.22)}} & 2.03 \scriptsize{\textcolor{HarderOrange}{(x1.49)}} & 133.02 \scriptsize{\textcolor{HarderOrange}{(x1.20)}} & 244.85 \scriptsize{\textcolor{HarderOrange}{(x1.24)}}\\ \dashline{2}{7}
& Vanilla (CoT) & \textbf{21.95} & \textbf{114.47} & - & \textbf{129.55} & -\\
& \cellcolor{lightgray!30}Ours (CoT) & 31.59 \scriptsize{\textcolor{HarderOrange}{(x1.44)}} & 135.08 \scriptsize{\textcolor{HarderOrange}{(x1.18)}} & - & 154.16 \scriptsize{\textcolor{HarderOrange}{(x1.19)}} & -\\ \midrule

\multirow{7}{*}{\shortstack{\textbf{Phi 3.5 Mini} \\ {\small Instruct (3.82B)}}} 
& Vanilla (Greedy) & \textbf{24.64} & \textbf{54.07} & \textbf{0.73} & \textbf{93.42} & \textbf{188.01}\\
& Beam Search & 85.77 \scriptsize{\textcolor{HarderRed}{(x3.48)}} & 177.82 \scriptsize{\textcolor{HarderRed}{(x3.29)}} & 2.06 \scriptsize{\textcolor{HarderRed}{(x2.84)}} & 270.75 \scriptsize{\textcolor{HarderRed}{(x2.90)}} & 636.49 \scriptsize{\textcolor{HarderRed}{(x3.39)}}\\
& \cellcolor{lightgray!30}Ours (Greedy) & \underline{33.57} \scriptsize{\textcolor{HarderOrange}{(x1.36)}} & \underline{57.22} \scriptsize{\textcolor{HarderOrange}{(x1.06)}} & \underline{0.97} \scriptsize{\textcolor{HarderOrange}{(x1.33)}} & \underline{113.25} \scriptsize{\textcolor{HarderOrange}{(x1.21)}} & \underline{255.51} \scriptsize{\textcolor{HarderOrange}{(x1.36)}}\\ \dashline{2}{7}
& Vanilla (Nucleus) & \textbf{26.49} & \textbf{56.63} & \textbf{0.73} & \textbf{94.00} & \textbf{188.27}\\
& \cellcolor{lightgray!30}Ours (Nucleus) & 37.07 \scriptsize{\textcolor{HarderOrange}{(x1.40)}} & 59.75 \scriptsize{\textcolor{HarderOrange}{(x1.06)}} & 0.97 \scriptsize{\textcolor{HarderOrange}{(x1.34)}} & 109.94 \scriptsize{\textcolor{HarderOrange}{(x1.17)}} & 262.50 \scriptsize{\textcolor{HarderOrange}{(x1.39)}}\\
\bottomrule
\end{tabular}

}
\caption{Average inference time in seconds for various models and methods across downstream datasets. Numbers in parentheses indicate the relative inference time compared to each model's Vanilla corresponding method. On average, \ours is x1.25, while beam search is x3.01, making it almost 2.5 times faster than beam search. This table complements Figure \ref{fig:inference_speed}. }
\label{tab:inf_speed}
\end{table*}
\section{Uncertainty in practice}
\label{adx:uncloc}

In Figure \ref{fig:uncloc}, we present an answer generated with \ours modified forward pass, where we highlight the steps with a \textproc{Shannon} entropy higher than the threshold $\theta$.
These uncertain steps happen right before a new decision is made.
We also notice that, unlike humans, the model does not encounter uncertainty while playing with numbers.
In this generation, \ours adjusted the top-1 predictions for five tokens before reframing.
The initial top-1 predictions, before reframing, are highlighted in blue in Figure \ref{fig:uncloc}.
Therefore, orange tokens without blue tokens are the ones where the initial prediction was already the current final token.

\begin{figure}[h]
    \centering
    \begin{tcolorbox}[boxrule=0pt, title=\textbf{Prompt:} Walter is collecting money for charity. First he collects \$500 from his neighbors. Then he collects \$1500 from a fund he set up online. His lawyer offers to donate three times as much as everyone else donates. Walter is going to tell him about his neighbors and the online fund until his friend gives him \$200 as well. How much is Walter's lawyer going to contribute?]
    \textbf{Answer}: \highlight{To} find out how much Walter's lawyer will contribute, we need to \highlight{\previoustok{calculate}find \previoustok{the}out \previoustok{the}how} much\highlight{ Walter has collected so} far\highlight{\previoustok{excluding}, and} then\highlight{ multiply it} by 3.\\
    Walter\highlight{ collected} \$500 from his neighbors\highlight{ and} \$1500 from the online fund.\\
    \highlight{\previoustok{Total}\$}500 + \$1500 = \$2000 \\
    \highlight{Then, his} friend gives him \$200.\highlight{ \\}\\
    \$2000 + \$200 = \$2200 \\
    \highlight{W}alter's lawyer\highlight{ will donate three} times as much as everyone else donates\highlight{. So, we multiply} \$2200 by 3\highlight{.\\}\\
    \$2200 * 3 = \$6600 \\
    \highlight{W}alter's lawyer will contribute \$6600.\\
    \end{tcolorbox}
    \caption{Answer to the given prompt generated using \ours. \highlight{Orange tokens} highlight additional forward steps (\ie tokens where uncertainty is higher than $\theta$). \colorbox{figBlue}{\st{Blue tokens}} represent the model's top-1 predictions prior to reframing.}
    \label{fig:uncloc}
\end{figure}

\section{Impact of the hesitation threshold}
\label{adx:theta_adx}

\begin{table*}[ht]
\centering
\scalebox{0.9}{
\begin{tabular}{c|cccccccc}
\toprule
\multicolumn{1}{c|}{\multirow{2}{*}{\textbf{Datasets}}} & \multicolumn{8}{c}{\textbf{Value of $\theta$}}\\
 & 0.6 & 0.8 & \textbf{1.0} & 1.2 & 1.4 & 1.6 & 1.8 & 2.0 \\ \midrule
CsQA & x1.03 & x1.02 & x1.01 & x1.00 & x1.00 & x1.00 & x1.00 & x1.00 \\
LAMBADA & x1.53 & x1.49 & x1.40 & x1.33 & x1.31 & x1.29 & x1.27 & x1.26 \\
\bottomrule
\end{tabular}
}
\caption{Relative inference time of \ours using LLaMA-3.1 Instruct 8B with different values of $\theta$ compared to the Vanilla method.}
\label{tab:theta_time}

\scalebox{0.9}{
\begin{tabular}{c|cccccccc}
\toprule
\multicolumn{1}{c|}{\multirow{2}{*}{\textbf{Datasets}}} & \multicolumn{8}{c}{\textbf{Value of $\theta$}}\\
 & 0.6 & 0.8 & \textbf{1.0} & 1.2 & 1.4 & 1.6 & 1.8 & 2.0 \\ \midrule
CsQA & +1.72 & +2.12 & +1.52 & +1.31 & +1.72 & +1.52 & +1.41 & +1.31 \\
LAMBADA & +5.12 & +5.21 & +5.16 & +5.12 & +5.08 & +5.02 & +4.95 & +4.85 \\
\bottomrule
\end{tabular}
}
\caption{Relative accuracy of \ours using LLaMA-3.1 Instruct 8B with different values of $\theta$ compared to the Vanilla method.}
\label{tab:theta_acc}
\end{table*}

Tuning the hyperparameter $\theta$ is beyond the scope of this paper.
However, we provide a preliminary study to analyze its impact. 
We evaluate two datasets with different $\theta$ values and report relative accuracy and inference time in Table \ref{tab:theta_time} and Table \ref{tab:theta_acc} respectively. 

As $\theta$ increases, inference speed improves thanks to fewer reframing steps---leading to reduced computation.
However, this comes at the cost of lower accuracy.
Conversely, a too small $\theta$ can also degrade accuracy.
This aligns with the discussion in Section \ref{sec:analysis} on multi-step reframing, where additional reframing may cause the model to ``overthink'' and lose track of improvements. 

This preliminary study is based on a limited set of tasks, and further investigation is needed to better understand the selection of $\theta$.
\newpage
\section{Extended Evaluation}
\label{adx:extended}

To further validate the robustness of our method, we conducted additional evaluations of \ours using the LM-Evaluation-Harness~\citep{eval-harness}.
This complements the original results in Section~\ref{sec:experiments} by offering a broader and more standardized assessment.

All experiments were conducted on the full datasets using a single LLaMA-3.1 8B Instruct model~\citep{dubey2024llama3herdmodels} in FP16 precision, without quantization, on an RTX A6000 GPU.
Each evaluation used six different random seeds to assess variability, except for MMLU Pro, which was run with one seed due to its computational cost.

CommonsenseQA and LAMBADA were evaluated using generation-based decoding, consistent with our initial setup, rather than LM-Eval’s default log-likelihood evaluation.
We report mean performance and standard deviation in Table~\ref{tab:add_results}.

\begin{table}[h]
    \centering
    \begin{tabular}{c|ccccc}
        \toprule
        \textbf{Method} & CsQA & GSM8K & LAMBADA & MMLU Pro & CNN/DM \\ \midrule
        Vanilla (Greedy) & 75.92 & 78.39 & 55.52 & 33.30 & 35.77 \\
        Ours (Greedy) & \textbf{76.19} \scriptsize{($\pm$ 0.39)} & \textbf{78.57} \scriptsize{($\pm$ 0.66)} & \textbf{56.62} \scriptsize{($\pm$ 0.26)} & \textbf{33.56} & 36.13 \scriptsize{($\pm$ 0.03)} \\
        \bottomrule
    \end{tabular}
    \caption{Extended evaluation results using LM-Evaluation-Harness. Scores reflect Exact Match for CommonsenseQA, MMLU Pro, and LAMBADA; accuracy for GSM8K; and ROUGE-1 for CNN/DM.}
    \label{tab:add_results}
\end{table}

These results confirm that \ours brings consistent gains across tasks.
However, they also highlight a key limitation: the introduction of randomness.
In particular, even when using greedy decoding, dropout-based reframing introduces stochasticity.
While the average performance improves, certain seeds can still negatively impact results.
This underscores the need for future work on reducing variance and improving robustness.
\newpage
\section{Prompts}
\label{adx:prompts}

This section presents the prompt for evaluating models on the different datasets.
``\emph{\{\dots\}}'' refers to the data extracted from the dataset itself for each example.

\begin{figure}[ht]
    \centering
    \begin{tcolorbox}[boxrule=0pt, title=\textbf{Multiple-choice Question} Prompt]
    \textbf{General Instructions}: Please read the multiple-choice question below carefully and select ONE of the listed options. Please write "The answer is LETTER".\\
    \textbf{Question}: \{\texttt{question}\}\\
    \textbf{Options}:\\
    \{\texttt{options}\}
    \end{tcolorbox}
    \caption{Multiple-choice Question prompt (CommonsenseQA and MMLU Pro).}
    \label{fig:prompt1}
\end{figure}

\begin{figure}[ht]
    \centering
    \begin{tcolorbox}[boxrule=0pt, title=\textbf{GSM8K}\ Prompt]
    \{\texttt{question}\}
    \end{tcolorbox}
    \caption{GSM8K prompt.}
    \label{fig:prompt2}
\end{figure}

\begin{figure}[ht]
    \centering
    \begin{tcolorbox}[boxrule=0pt, title=\textbf{LAMBADA}\ Prompt]
    You have to predict the next word of the sentence given the context. Your answer should be a single word.\\
    \textbf{Context}:\\
    \{\texttt{context (minus one sentence)}\}\\
    \textbf{Sentence to continue}:\\
    \{\texttt{last sentence (minus one word)}\}
    \end{tcolorbox}
    \caption{LAMBADA prompt.}
    \label{fig:prompt3}
\end{figure}

\begin{figure}[ht]
    \centering
    \begin{tcolorbox}[boxrule=0pt, title=\textbf{CNN/DM}\ Prompt]
    Summarize the given article. Do not output something else than the summary.\\
    \textbf{Article}:\\
    \{\texttt{article}\}
    \end{tcolorbox}
    \caption{CNN/DailyMail prompt.}
    \label{fig:prompt4}
\end{figure}

\end{document}